\documentclass{article}

\usepackage{wrapfig}
\usepackage[final]{corl_2019} % Uncomment for the camera-ready ``final'' version
\usepackage[utf8]{inputenc} % allow utf-8 input
\usepackage[T1]{fontenc}    % use 8-bit T1 fonts
\usepackage{url}            % simple URL typesetting
\usepackage{booktabs}       % professional-quality tables
\usepackage{amsfonts}       % blackboard math symbols
\usepackage{amsmath}
\usepackage{amssymb}
\usepackage{nicefrac}       % compact symbols for 1/2, etc.
\usepackage{microtype}      % microtypography
\usepackage{lib}

\usepackage{graphicx}
\usepackage{subfig}
\usepackage[font=footnotesize,labelfont=bf]{caption} % to make caption text small, check if nips likes it.
\usepackage[title]{appendix}

\newenvironment{tightlist}%
{\begin{list}{$\bullet$}{%
    \setlength{\topsep}{0in}
    \setlength{\partopsep}{0in}
    \setlength{\itemsep}{2mm}
    \setlength{\parsep}{0in}
    \setlength{\leftmargin}{2em}
    \setlength{\rightmargin}{0in}
    \setlength{\itemindent}{0.1in}
}
}%
{\end{list}
}

\begin{document}
% \title{Explicitly structured Prediction for Manipulation}
\title{Object-centric Forward Modeling \\ for  Model Predictive Control}
% \title{Learning Compositional Forward Modeling \\ for Model Predictive Control}
% The \author macro works with any number of authors. There are two
% commands used to separate the names and addresses of multiple
% authors: \And and \AND.
%
% Using \And between authors leaves it to LaTeX to determine where to
% break the lines. Using \AND forces a line break at that point. So,
% if LaTeX puts 3 of 4 authors names on the first line, and the last
% on the second line, try using \AND instead of \And before the third
% author name.

% NOTE: authors will be visible only in the camera-ready (ie, when using the option 'final'). 
% 	For the initial submission the authors will be anonymized.

\author{Yufei Ye\textsuperscript{1}  \qquad Dhiraj Gandhi \textsuperscript{2} \qquad Abhinav Gupta\textsuperscript{12} \qquad Shubham Tulsiani\textsuperscript{2} \\
% \vspace{-2mm}
\textsuperscript{1}Carnegie Mellon University  \qquad \textsuperscript{2}Facebook AI Research \\
{\tt \small yufeiy2@cs.cmu.edu \qquad \{dhirajgandhi,gabhinav,shubhtuls\}@fb.com}
\\
{\tt \small \href{https://judyye.github.io/ocmpc/}{https://judyye.github.io/ocmpc/}}
}

% \author{
%   Yufei Ye \\
%   \texttt{yufeiy2@cs.cmu.edu} \\
% \And  Dhiraj Gandhi  \\
%   \texttt{dhirajgandhi@fb.com} \\
% \AND    Abhinav Gupta* \\
%   \texttt{abhinavg@cs.cmu.edu} \\
% \And    Shubham Tulsiani* \\
%   \texttt{shubhtuls@fb.com} \\
% }
\maketitle

\vspace{-5mm}
\begin{abstract}
% We want to do better control with explicitly structured prediction module.
We present an approach to learn an object-centric forward model, and show that this allows us to plan for sequences of actions to achieve distant desired goals. We propose to model a scene as a collection of objects, each with an explicit spatial location and implicit visual feature, and learn to model the effects of actions using random interaction data. Our model allows capturing the robot-object and object-object interactions, and leads to more sample-efficient and accurate predictions. We show that this learned model can be leveraged to search for action sequences that lead to desired goal configurations, and that in conjunction with a learned correction module, this allows for robust closed loop execution. We present experiments both in simulation and the real world, and show that our approach improves over alternate implicit or pixel-space forward models. Please see our \href{https://judyye.github.io/ocmpc/}{project page} for result videos.
\end{abstract}

% \keywords{Compositionality, Forward Model,  Predictive Control} 
\section{Introduction}
% Consider the configuration of the objects shown in Figure 1. How can a robot, by simply pushing things around, 
\begin{wrapfigure}{r}{0.34\textwidth}
  \begin{center}
\vspace{-2em}
    \includegraphics[width=0.34\textwidth]{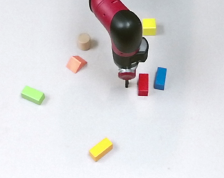}
  \end{center}
  \caption{What happens when the robot arm moves right?}
  \vspace{-1em}
\end{wrapfigure}
What will happen if the robot shown in Fig 1. moves its hand to the right by a few inches? We can all easily infer that this will result in the red block moving right, and possibly hitting the blue one, which would then also move. This ability to perform \emph{forward modeling} \ie predicting the effect of one's actions is a cornerstone of intelligent behaviour -- from realizing that turning the knob opens a door, to understanding that falling off from a height can lead to an injury. Not only does this allow us to judge actions based on their immediate consequences, it also enables us to reason about \emph{sequences of actions} needed to achieve desired goals. As an illustration, let us again consider Fig 1, but now try to imagine how we can get the red block to the right end of the table, but without disturbing the blue object. We know that this can be achieved by first pushing the object up, then towards the right, and then back down. This seemingly simple judgment is actually quite remarkable. In addition to understanding that the naive solution of simply pushing right would not succeed, we also could find, among myriad other possibilities, the sequence of actions that would -- by chaining together our understanding of each individual action to understand the effect of the collective. In this work, our goal is to build agents that can exhibit similar capabilities \ie given an image, understand the consequences of their actions, and leverage this ability to achieve distant goals.

% \todo{we need a Fig 1. with few objects and real robot arm}
% para 1: given a fig 1, we can predict what happens. this also allows us to plan what to do given an end goal. say this is what we do in this work.

% This insight of using a `forward model' to find action sequences that lead to desired outcomes is a well-established one in the field of AI, dating back to methods handling discrete search spaces with known `successor functions' to solve games such as chess~\cite{}. These approaches have since been successfully adapted to robotic manipulation tasks~\cite{} for scenarios where the state of the system can be easily represented, and the effect of actions analytically obtained via knowledge of the underlying physics.
This insight of using a `forward model' to find action sequences that lead to desired outcomes is a classical one in the field of AI, and has been successfully adapted to robotic manipulation tasks~\cite{lynch1996stable,zhou2016convex} for scenarios where the state of the system, \eg shape, pose, mass \etc of objects, can be easily represented, and the effect of actions analytically obtained.
While this explicit state representation can allow efficient and accurate planning, understanding the state of the system from visual observations or analytically modeling the dynamics is not always possible. Some recent approaches~\cite{agrawal2016learning,ebert2017self} therefore propose to learn a forward model over various alternate representations \eg implicit features, or pixel space.
However, we argue that using these implicit or pixel based representations for forward modeling discards the knowledge about the structure of the world, thereby making them less efficient or accurate. When we think of the scene in Fig 1, and the effects of our actions, we naturally think of the different blocks, and their interactions with each other or the robot. Towards learning forward models that have similar inductive biases, 
%Although this allows inference and planning from visual input, these representations do not leverage the structure of the problem, thereby making them less efficient or accurate.
we propose to use a semi-implicit representation -- explicitly encoding that a scene comprises of different \emph{entities}, but having an implicit representation to capture the appearance of each entity. 

Concretely, we represent each object using its spatial location (in image space) and an implicit feature that is descriptive of its appearance (and can implicitly capture transforms like rotations). We build on this object-centric scene representation and present an approach that learns to model the effect of actions in a scene via predicting the change in representation of the objects present while allowing for interactions between the objects and the robot, as well as among the objects.  This object-centric forward model allows us to capture several desirable inductive biases that help in learning more efficient and accurately models -- a scene comprises of spatial objects, actions can affect these objects, and the objects can, in turn, affect each other. We show we can leverage our learned model to search for a sequence of actions that would allow us to reach a desired scene configuration from the current input image. However, as the forward model is not perfect, we additionally propose to use a `refinement' module that can re-estimate the scene configuration in the context of the observed image. This allows us to robustly act in a closed loop manner to achieve desired goal configurations, and we show that our approach improves over previous pixel-space or implicit forward models.

% A key question for any such approach is regarding the form of the representation over which the forward model is learned, and previous approaches have explored various alternatives -- from explicit physical state based~\cite{} to implicit feature based representations~\cite{}. While explicit state representations can allow one to fully specify a physical system and disentangle factors of variation, 
% We instead propose to leverage a semi-implicit representation -- explicitly encoding that a scene comprises of different `entities', but having an implicit representation for each.

% While explicit state representations can allow one to fully specify a physical system  
%para 2: talk of why `entities' is a good idea

%para 3: previous works have focused on entity-less models, we instead decouple entity location and implicit feature

%para 4: this factorization lets us learn good forward models, which we can use for planning. talk of the `refinement' step. conclude by saying we get great results.

\section{Related Work}
%\vspace{1mm}
\noindent \textbf{Learning Video Prediction.}
While forward modeling aims to predict the future conditioned on an action, a related task in computer vision community is that of simply predicting the future, independent of any action. Several approaches have attempted to predict videos in pixel space ~\cite{vondrick2016generating, walker2016uncertain, mathieu2015deep}.  Instead of directly regressing to pixels, alternate flow-based prediction models have also shown promising results~\cite{finn2016unsupervised, xue2016visual}.  However, these can typically only handle small motions between frames, and need a large number of samples to overcome this inductive bias. Most related to our work is the approach by Ye \etal~\cite{cvp}, which also pursues prediction in an object-centric space, and in this work we show these can be extended to action-conditioned prediction and planning.

\vspace{1mm}
\noindent \textbf{Predictive Forward Models for Acting and Planning.} 
In the robotics community, learned forward models have been used for a plethora of tasks \eg leveraging forward models for exploration~\cite{pathak2017curiosity, oh2015action, gandhi2017learning}, or to learn a task policy~\cite{wahlstrom2015pixels, hafner2018learning, deisenroth2011pilco}. More related to ours, some approaches~\cite{agrawal2016learning, nair2017combining} jointly learn a forward and inverse model, where the latter is regularized by the former and can be used to greedily output actions given current observation and a short-term goal. We adopt the philosophy for some recent methods~\cite{ebert2018robustness, watter2015embed} that also tackle longer horizon tasks, by training a forward model and then using a planner to  generate actions. However, these methods still face difficulty in handling large change per action or a large number of actions. We overcome these limitations by leveraging object-centric representations for forward modeling and planning. 

\vspace{1mm}
\noindent \textbf{Structured Models for Physical Interactions.} 
Rather than predicting in implicit or pixel-space representation, a line of work with a motivation similar to ours, models physics by explicitly modeling the state transitions, using known~\cite{fragkiadaki2015learning} or predicted~\cite{wu2016physics,xu2019densephysnet} physical properties. However, for generic manipulation tasks  in the real world, the dynamics and physical properties cannot easily be captured analytically. Recent learning-based works~\cite{pushnet} overcome this in a data-driven manner, and show impressive results for forward modeling and planning with previously unseen, but isolated objects. Towards handling more generic scenarios, some approaches leverage graph neural networks to reason about the interaction between objects~\cite{battaglia2018relational, watters2017visual, kipf2018neural, chang2016compositional}, but typically apply their methods to simpler scenarios that do not involve robotic manipulation and where the state can be estimated.
Janner \etal~\cite{janner2018reasoning} show that such compositional forward models can be applied for tasks like block stacking, but learn these for predefined high-level action primitives. In contrast, our work targets forward modeling for low-level continuous control, where a long sequence of actions is required to achieve a goal.

\section{Approach}
\begin{figure}
    \centering
    \includegraphics[width=\linewidth]{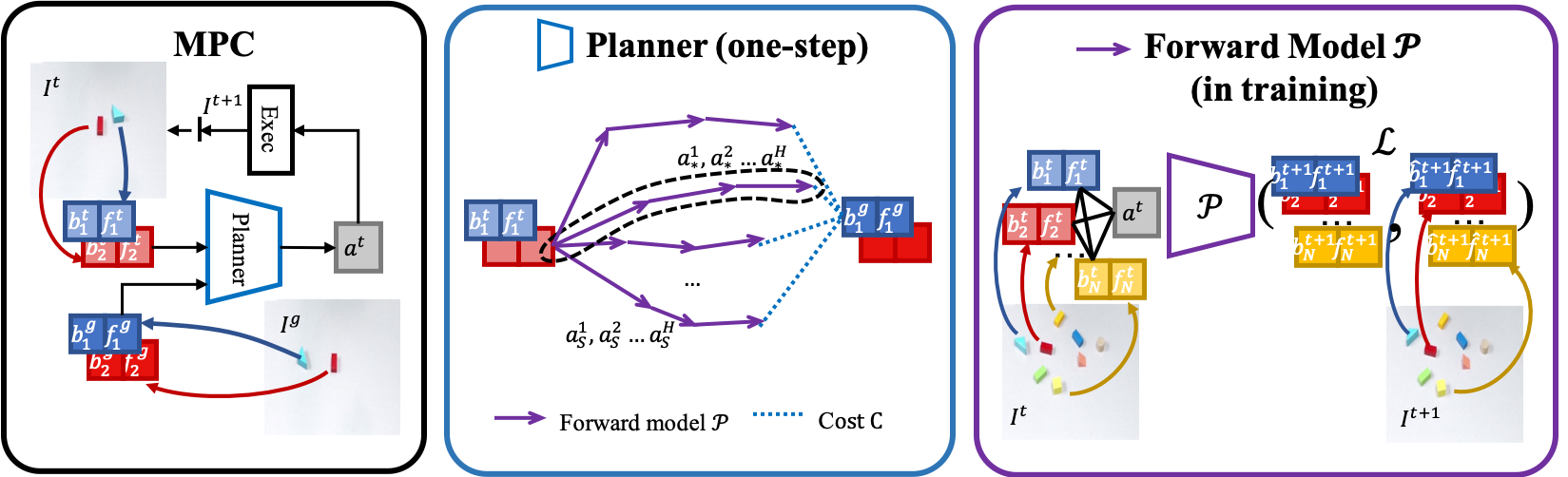} \vspace{1mm}
    \caption{Left: We demonstrate MPC in testing time. Given a goal and initial configuration, the planner takes as input the object-centric representation (section \ref{sec:representation}) and outputs an action to execute. Then, a new observation is obtained to repeat the loop. Middle: Inside the planner, several action sequences are sampled and unrolled by the forward model (section \ref{sec:forward}). The best sequence with respect to the cost  is selected, among which only the first action is executed (section \ref{sec:planning}). 
    Right: The forward model $\mathcal{P}$ takes as input a representation of a scene with an action and predicts the next step. It is supervised by the ground-truth representation of the future. 
    % \todo{judy / dhiraj: caption + fig, see R3. ST's comment: we need to change images here} We learn an object-centric forward model $\mathcal{P}$ that takes as input a representation of a scene in the form of explicit spatial location of each object (\todo{DG: could you add $b_n^t, f_n^t$ in figure? }darker color) and associated implicit visual features (lighter color). Given an action, our model outputs a prediction in the form of a similar object-centric representation. This learned model allows us to search for action sequences that lead to desired goal configurations.
    }
    \label{fig:my_label}
\end{figure}

Given an image depicting multiple objects on a table, and a goal image indicating the desired configuration, we aim to plan a sequence of pushing actions $a^{1:T}$ such that executing them leads to the goal. To search for an optimal action sequence, a forward model is essential to hallucinate the future configurations that would result from any such sequence. Our insight is that to manipulate in a complicated environment with multiple objects, object-level abstraction is crucial to make long-horizon predictions and plans. We propose to use an object-centric forward model that can predict the effect of actions via implicitly modeling the interactions among the objects and the robot. While the learned model allows planning using the object-centric representation, our estimate of the objects' locations after a performed action is not perfect and needs to be re-estimated for closed loop execution. We therefore also propose to learn a refinement module to correct our predicted representation using the updated observation.

% We follow the framework of Model Predictive Control (MPC \cite{mayne2014model}). In MPC, a forward model predicts the future state $x^t$ given the current state $x^0$ and action sequence $a^{1:t}$. The MPC planner minimizes the distance (cost) between predicted state $x^t$ and $x^g$ with respect to action sequence $a^{1:H}$, where $H$ is the planning horizon. Only the first action in the planning sequence is executed. Then the sequence is replanned with feedback observations. In our work,  \todo{one line on refinement here}

\vspace{-1mm}
\subsection{Object-Centric Representation.}
\vspace{-1mm}
\label{sec:representation}
Given an observation in the form of an image, we aim to predict, in some representation, the effect of actions, and then plan towards desired goals using this learned model. What the form of this representation should be is an open research question, but it should be  efficient to learn both prediction and planning with.
%we aim to propose a representation that is easy to predict and to plan. It is an open research question about the space in which predictions should be  made. 
Our insight is to explicitly leverage the fact that multiple distinct objects are present in typical scenes, and they can  naturally be represented as `where' and `what' \ie their location and visual description. We operationalize this insight into our representation.

Concretely, given an observed image $I^t$ and (known/predicted) location of N objects $\{b_n^t\}_{n=1}^{N}$ in the image, we use an object-level representation as $\{x_n^t\}_{n=1}^{N}$. Each object is represented as   $x_n^t \equiv (b_n^t, f_n^t)$, where $b_n^t$ is the  observed/predicted location and $f_n^t$ is the implicit visual feature of that object which encodes rotation, color, geometry, etc. The location $b_n^t$ is simply the $xy$-coordinate in image space.  $f_n^t$ is a feature extracted from a fixed sizes window centered on $b_n^t$, extracted by a neural network with ResNet-18 ~\cite{he2016deep} as backbone. 
%The feature  is 128. 

% We argue it is easy to access rough location with an off-the-shelf visual detector.

\vspace{-1mm}
\subsection{Object-Centric Forward Model}
\vspace{-1mm}
\label{sec:forward}
Given the current object-centric descriptor  $\{x_n^t\}_{n=1}^N = \{(b_n^t, f_n^t)\}_{n=1}^N$ of current time $t$, and an action about to execute $a^{t+1}$, the forward model $\mathcal{P}$ predicts the representation $\{x_n^{t+1}\}$ for each object at the next time step $t+1$, i.e.\ $\{x_n^{t+1}\} \equiv \mathcal{P}(\{x_n^t\}, a^{t+1})$. 
% Given the current object-centric descriptor $\{x_n^t\}_{n=1}^N$, We train a one-step forward model $\mathcal{P}$ to predict representations at time $t+1$ conditioned on the aciotn $a^{t+1}$. More formally, a forward model aims to output the future representation $\{x_n^{t+1}\} = \mathcal{P}(\{x_n^t\} \bigcup a^{t+1})$.
To predict the effect of a longer sequence of actions $a^{t:t+H}$, we can simply just  apply the forward model iteratively $H$ times to hallucinate representation at time $t+H$.
% Given a descriptor $\{x_n^t\}_{n=1}^N\equiv \{(b_n^t, f_n^t)\}_{n=1}^N$ of current time $t$, and an action about to execute $a^{t+1}$, the forward model predicts the representation $\{x_n^{t+1}\}$ for each object at the next time step $t+1$, i.e.\ $\{x_n^{t+1}\} = \mathcal{P}(\{x_n^t\}, a^{t+1})$. 

To allow modeling the interaction among robot and objects, the forward model $\mathcal{P}$ is implemented as an instance of Interaction Network\cite{battaglia2016interaction}. In general, the network takes in a graph $(V, E)$, where each node is associated with a vector. The network learns to output a new representation for each node by iterative message passing and aggregation. The message passing process is inherently analogous to physical interactions. % \todo{concrete example?} 
To allow for robot-object and object-object interaction, besides each object represented as a node, the action of the gripper is added as an additional node, with the features being a learned embedding of the action $a^{t+1}$.
%\todo{change notation to use consecutive frames?}
In addition to the predictor $\mathcal{P}$, we also train a decoder $\mathcal{D}$ to further  regularize features to encode meaningful visual information. Similar to ~\cite{cvp},
the decoder takes in $\{x_n^t\}$ and decodes to pixels.

To train the forward model, we collect training data in the form of triplets $( I^t, a^{t+1}, I^{t+1})$, where $ I^t$ denotes observed images,  In addition, we also require location of each object at every time step $\{\hat b_n^t\}$ and the correspondence of those objects across time.  We argue these annotations (with some possible noise) can be obtained using an off-the-shelf visual detector, as we demonstrate on real robot data in section \ref{sec:exp:sawyer}. We supervise the model using a combination of two losses -- a reconstruction loss and a prediction loss. The reconstruction loss forces the features to encode meaningful visual information (and prevent trivial solutions), and the prediction loss encourages the forward model to predict plausibly both in feature space and in pixel space.
\begin{align*}
    L_{recon} &= \| \mathcal{D}(\{\hat x_n^t\}) - I^{t}\|_1 \\
    L_{pred} &= \| \mathcal{D}(\{ x_n^{t+1}\}) -  I^{t+1}\|_1 + \| \mathcal{P}(\{ \hat x_n^{t}\}, a^{t+1}), \{\hat x_n^{t+1}\}\|^2_2
\end{align*}
where $\hat x^t_n$ represents features extracted from $I^{t}$ at the ground-truth object locations.
% This section starts with introducing our object-centric representation, followed by describing how to predict the change of the representation given a pushing action. \todo{the gripper is also a node, but when should this part come in? } %  how to train the prediction model in self-supervised way. 

\vspace{-1mm}
\subsection{Planning Via Forward Model}
\vspace{-1mm}
\label{sec:planning}
Given this learned forward model $\mathcal{P}$, we can leverage it to find action sequences that lead to a desired goal. Specifically, given the goal $\{x_n^g\}$ and current state $\{x_n^0\}$, we generate an action trajectory $a^{1:T}$ such that executing them would lead towards the goal configuration. 

We optimize the trajectory by a sample-based optimizer -- cross entropy method (CEM) \cite{rubinstein1999cross}. In CEM, at every iteration, it draws $S$ trajectories of length $H$ from a Gaussian distribution, where $H$ is the planning horizon. The forward model evaluates those sequences by computing the distance of the predicted state $\{x^{H}_n\}$ to the goal configuration. The best $K$ samples are then selected with which a new Gaussian distribution is refit.
The function to evaluate distance of two states / cost of actions is:
$$
C = \sum_{n=1}^N (\|b_n^H, b_n^g\|_2^2 + \lambda \|f_n^H, f_n^g \|^2_2)
$$

After $\tau$ iteration of optimization, the trajectory leading to the lowest distance to the goal is returned. Instead of executing the whole sequence of length $H$, only the first step is actually applied. Then we observe the feedback and re-plan the next action.
%, following the framework of Model Predictive Control (MPC).

% Note that in contrast with related works, our cost function only computes the distance between the goal and the $H$-th state in the horizon to encourage non-greedy plans. It is possible because prediction of the forward model still remain very sharp and plausible after unrolling multiple times. 
In our experiment, $S=200, K=10, \lambda=100, \tau=3, H=5$. We sample trajectories in continuous velocity space and upper-bound the magnitude.
% \todo{shall we highlight that we can only train with one-step?? }
% With the predictor, how to plan with the 
% \todo{Not sure about contribution of this part.... I almost follow chelsea's setup but remove some seemingly hacking to a more standard? MPC. (1) they repeat sampled actions 3 times at every step, and execute every the first 3 actions. (2) reduce the variance for action $a_{2:\tau-1}$ for next planning iteration.}

\vspace{-1mm}
\subsection{Robust Closed Loop Control via Correction Modelling}
\vspace{-1mm}
\label{sec:correction}
We saw that given a current representation $\{x^t_n\}$ and the desired goal configuration $\{x^g_n\}$, we can generate a sequence of action $a^{t+1:t+1+H}$, among which the first action $a^{t+1}$ is then executed at every time step, after which we replan. However, as we do not assume access to ground-truth object locations at intermediate steps, it is not obvious what the new `current' representation \ie $\{x^{t+1}_n\}$ should be for this re-planning. One option is to simply use the predicted representation $\{x^{t+1}_n\} \equiv \mathcal{P}(\{ x_n^{t}\}, a^{t+1})$, but this leads to an open loop controller where we do not update our estimates based on the new observed image $\hat I ^{t+1}$ that we observe after our action. As our prediction model is not perfect, such a predicted representation would then quickly drift, making robust execution of long-term plans difficult.

% Before replanning a new sequence, we observed a new image $\hat I ^{t+1}$ which could refine our prediction. Had been no refinement, the predicted location will slowly drift and lose track of the object since the forward model cannot be perfect.  Hence, how should it update the location given observed raw pixels?  

To solve this problem, we propose to additionally learn a correction model $\mathcal{C}$ that can update the predicted location based on the new observation image $\hat I ^{t+1}$. Denote $I[b]$ as the region cropped on image $I$ specified by the location $b$. Given the initial crop $\hat I^0 [ \hat b_n^0]$ to visually describe the object being tracked, and the predicted location cropped on the new observed image
$\hat I^{t+1}[b_n^{t+1}]$,  it regresses the  residual $\Delta b_n^{t+1}$ to refine, such that $b_n^{t+1} + \Delta b_n^{t+1} $ approximates $\hat b_n^{t+1}$ and re-centers the cropped region to objects. We train this model using random jitters around the ground truth boxes on the same training data used to learn the forward model. %\todo{improve notation, explain bit better.}
%More specifically, the correction model first embed regions to feature space, concatenate them, and perform regression on top of that, i.e.\ $\Delta b_n^{t+1} \equiv \mathcal{C}(\hat I^{t+1}, \{ b_n^{t+1}\}, \hat I^0, \{\hat b_n^0\})$.

% In real life, neither prediction nor location estimation can be perfect. When testing with only RGB image, we also cannot not estimate precise location at each time step. The prediction cannot be perfect, it might drift off the object thus lose track when we unroll predictor multiple times. \cite{registration? and retrying}. 

\section{Experiments}
Our goal is to demonstrate that our learned object-centric forward model allows better planning compared to alternatives. To this end, we evaluate our method under both synthetic and real-world settings, and observe qualitative and quantitative improvements over previous approaches.

\vspace{-1mm}
\subsection{Experimental Setup}
\vspace{-1mm}
\begin{figure}
    \centering
    \includegraphics[width=\linewidth]{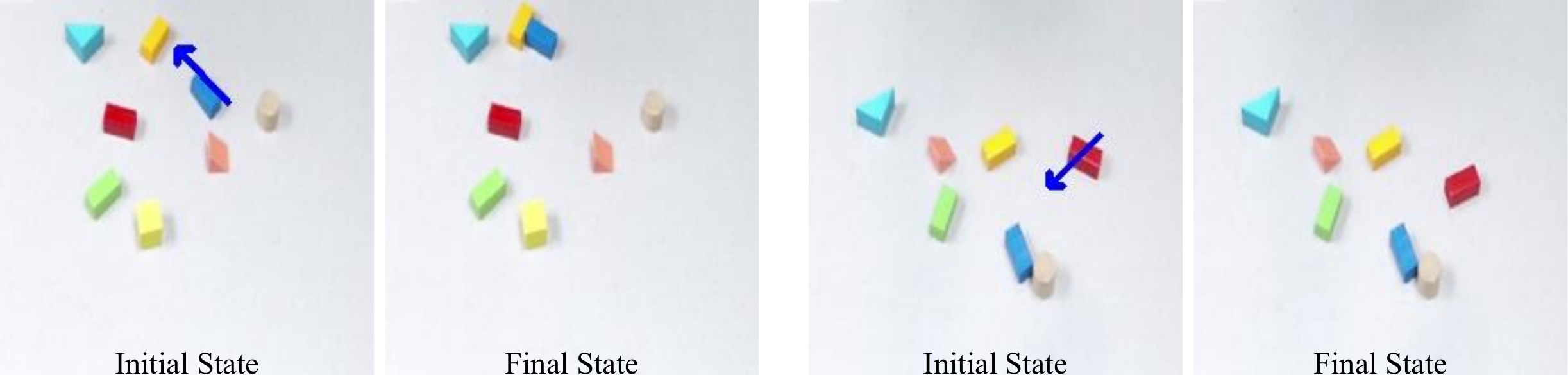}
    \caption{Examples of initial state and final state images taken for the push action in real world. The arrow depicts the direction and magnitude of the push action.}
    \label{fig:eral_data}
    \vspace{-1.5em} 
\end{figure}

\noindent \textbf{Collecting Training Dataset.} We work on two pushing datasets, a synthetic environment in MuJoCo \cite{todorov2012mujoco} and on a real Sawyer robot. To collect training data, multiple objects are scattered on a table. The robot performs random pushes and records the observation before and after. The push action $a^t$ is represented as the starting and ending  position of the end effector in world coordinate.   

In the synthetic dataset, we generate 10k videos of pushing two randomly composed L-shape objects on the table. Each video is of length 60 (600k pushes in total) and motion between frames is relatively small. To train our prediction model, we extract the ground-truth locations from the MuJoCo  state.

To collect real-world data, we generate 5k random pushes (10k images), where the length of each push is relatively large. As a result, in some of these actions, objects can undergo large motion (Figure \ref{fig:eral_data}).  To obtain the location and correspondence of objects in training set, we manually annotated around 30 images to train a segmentation network~\cite{ronneberger2015u}. The location is assumed to be the center of the corresponding mask. All of the data collected for the experiment is publicly available at \href{https://github.com/CMURobotAndVisualLearningLab/Pushing-Dataset.git}{data link}.

\noindent \textbf{Evaluation Setup.} In both synthetic and the real world, the test set is split into two subsets with one object and two objects, respectively.  For quantitative evaluation, we evaluate our model and baselines in simulation, using the distance of objects to the goal position as the metric.  In the simulated test set, the distance of initial configuration to the goal is 15 times larger than the length of a single push. The locations are only available to models for the initial and goal configuration, but not at the intermediate steps. In those intermediate steps, only new images are observed and state information is updated and estimated by the models themselves. In the real robot setting, we manually create some interesting cases for qualitative comparisons, such as manipulating novel objects and when the robot has to predict interactions to avoid other objects. % This setting is similar to  \cite{finn}.

\noindent \textbf{Baselines.} We compare our approach of the object-centric forward modelling with the following baselines and their variants:

\begin{tightlist} 
    \item Implicit forward model \cite{agrawal2016learning} (\textbf{Imp-Inv}/\textbf{Imp-Plan}): We follow Agrawal \etal~\cite{agrawal2016learning} and learn a forward model in a feature space where the entire frame is encoded as one implicit feature. Imp-Inv generates actions greedily by the inverse model which takes in current and goal feature; Imp-Plan plans action sequence in the learned representation space.  
    % An inverse model is simultaneously trained to take in current and goal feature and outputs the action. Following ~\cite{agrawal2016learning}, we do not plan a sequence of action but rather iteratively query the inverse model given current and goal image.
    % \item Implicit forward model with reconstruction loss for planning (\textbf{Imp-Plan}): The learned representation is used by the planner rather than a greedy inverse model. We also add a reconstruction loss in pixels to encourage more informative feature space.
    \item Flow-based prediction model \cite{ebert2017self} (\textbf{Flow}/\textbf{Flow-GT}): We follow Ebert \etal~\cite{ebert2017self} and learn to predict transformation kernel to reconstruct future frame.  In planning, the predicted transformations are applied to designated pixels (location) to estimate  their motion. Two flow baselines update the state information by maintaining the probability maps of designated pixels as in previous work. During training, Flow-GT is additionally leverages  known object locations during training by supervising the desired transform for the object centre locations.
    \item \textbf{Analytic} baseline: If the exact center of mass is known at each step, a straightforward solution is to greedily push towards goal position. This baseline assumes a naive forward model -- the change of location at the next step will be same as  the change of gripper  position.
\end{tightlist}

\vspace{-1mm}
\subsection{Experiment with Synthetic Environment}
\vspace{-1mm}
\begin{figure}
    \centering
    \includegraphics[height=1.3in]{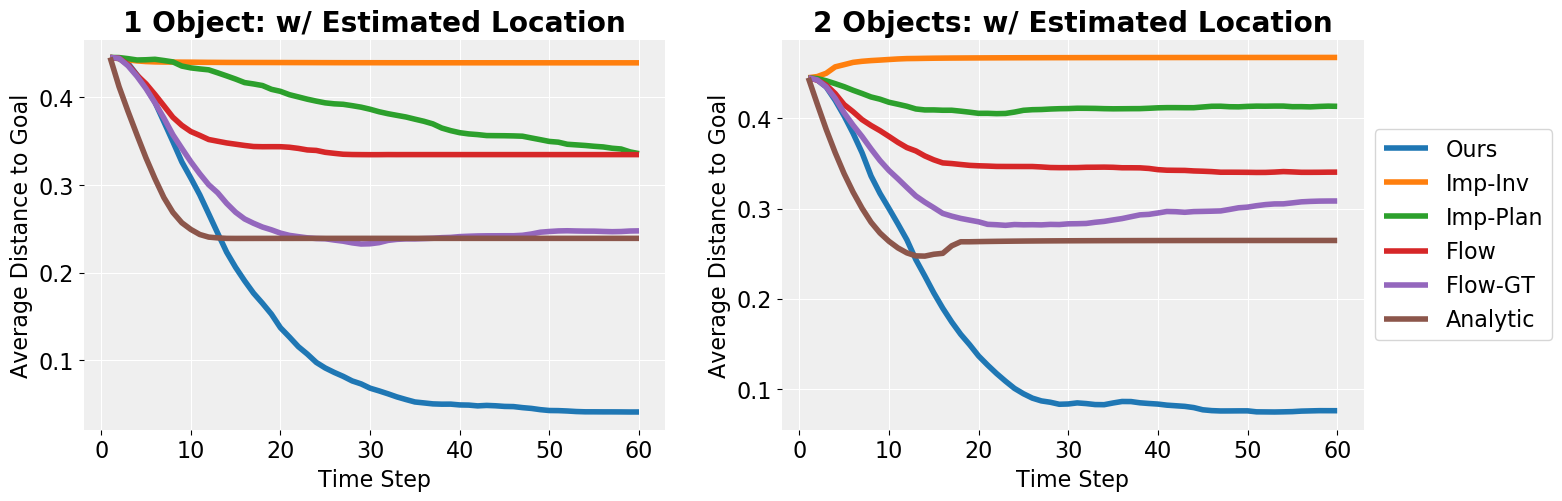}
    \caption{Quantitative results on synthetic dataset: Plot shows the distance between the current configuration and the goal configuration at different time steps with 1 object(left) and 2 objects (right) in the scene. }
    % \vspace{-2.5em}
    \label{fig:exp:est}
\end{figure}
\begin{figure}
    \centering
    \includegraphics[width=\linewidth]{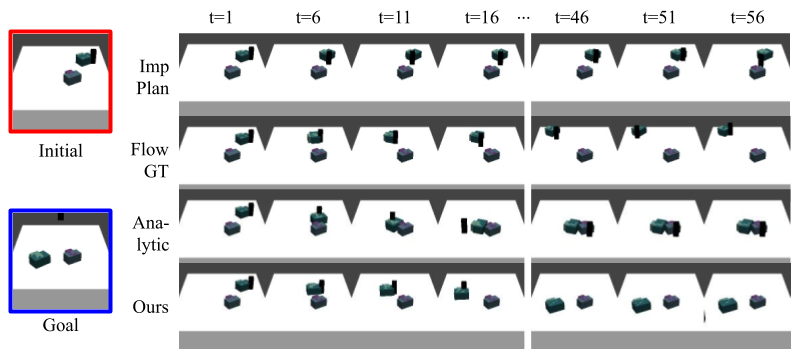}
    \caption{Visualizing an executed action sequence in simulation: Given the initial configuration (in red box) and the goal configuration (in blue box), figure shows the effect of the action predicted by various methods at different time steps. Please refer to the appendix to view all baselines.}
    \vspace{-1.5em}
    \label{fig:exp:est_vis}
\end{figure}
% In testing, the location is only given for the initial observation and the goal configuration.  In the intermediate  steps during execution, the model should maintain and update the state by itself. 

We measure the performance across methods by analyzing the average distance of objects from their goal positions. We plot the average distance over time between the current location and the goal in world coordinate in Figure \ref{fig:exp:est}. We find the the the `Imp-Inv' fails to generalize to scenarios when the distance of goal and current observation is much ($\times 15$) farther than that in training set, thereby showing the importance of planning rather than using a one-step inverse model. The `Imp-Plan' baseline degenerates significantly for 2 blocks, suggesting a single feature cannot encode the whole scene very well. 
Flow baseline works much better than Imp-Plan because the motion space is more tractable compared to implicit feature space of frames. Its performance further improves by leveraging location information during training, as seen by the `Flow-GT' curve. 
However, using our object-centric model for planning further improves over these baselines as shown in Figure \ref{fig:exp:est}. 

Figure \ref{fig:exp:est_vis} showcases an interesting example where one block needs to be pushed around the other to reach the goal. In this particular case, learning-based alternatives fail to search a plausible plan. The analytic baseline performs well at the beginning with simple dynamics but loses track of the object when the block collides with the other. In contrast, our approach carries out the correct action sequence and manages to reach the goal, demonstrating that we can reason about interaction among objects.
For more qualitative results, please refer to our website. 

\vspace{-1mm}
\subsection{Experiments with Real Robot}
\begin{figure}
    \centering
    \includegraphics[width=0.9\linewidth]{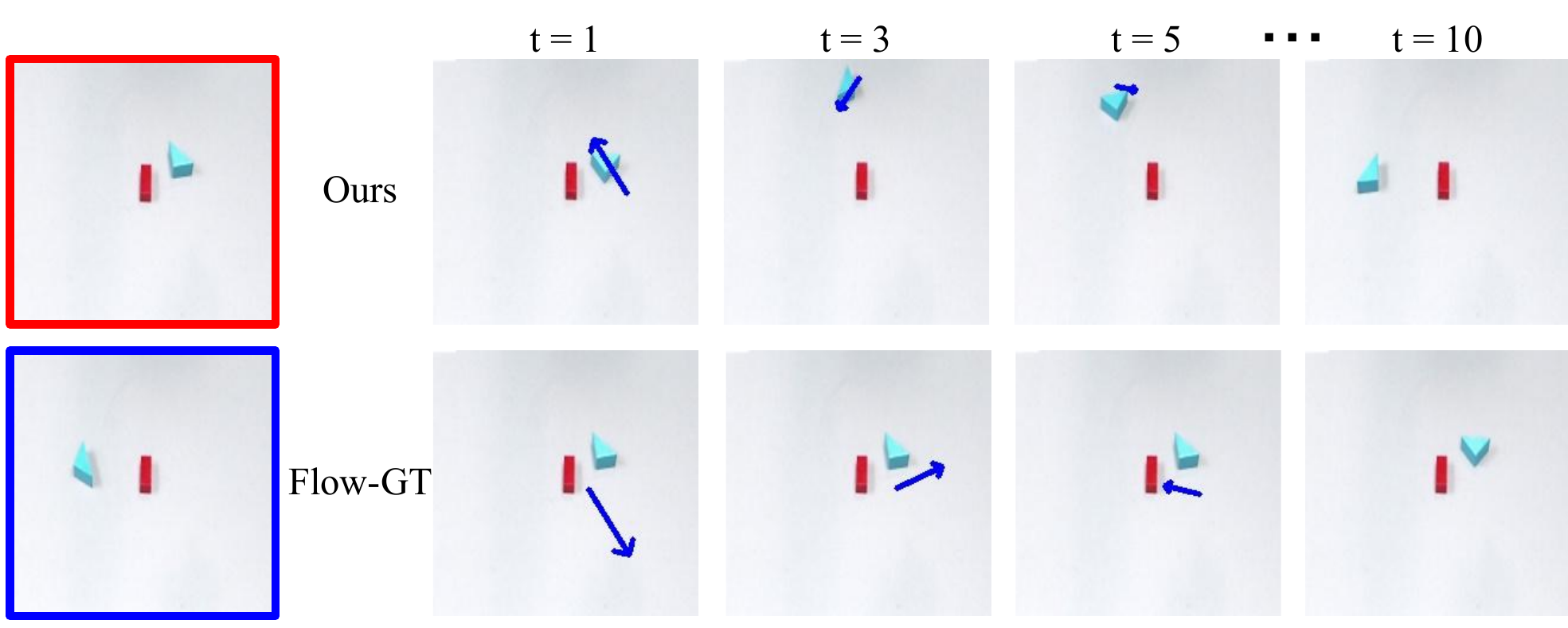}
    \caption{Visualizing an executed action sequence in real world: Given the initial configuration (in red box) and the goal configuration (in blue box), blue arrow shows the sequence of action taken by the robot.}
    \label{fig:exp:real_vis}
\end{figure}

\vspace{-1mm}
\label{sec:exp:sawyer}
In the real robot setting, we compare our model with the best performing baseline based on the synthetic results \ie `Flow-GT'. Figure \ref{fig:exp:real_vis} shows a qualitative result with two blocks. Similar to the example in synthetic data, to push the blue block to the goal position, our model manages to carry out a plan which avoids the red block in between. In contrast, Flow-GT generates relatively random actions, probably because the large motions that can result from a single push are difficult to model.
%We ablate the prediction model in Section \ref{sec:exp:ablation}.
We present additional results in the appendix, and also show that our model can generalize to novel objects by training with simple blocks.

\vspace{-1mm}
\subsection{Ablations}
\vspace{-1mm}
\label{sec:exp:ablation}

% \noindent \textbf{}
\begin{figure}
% 	\vspace{-1em}
    \centering
    \includegraphics[height=1.25in]{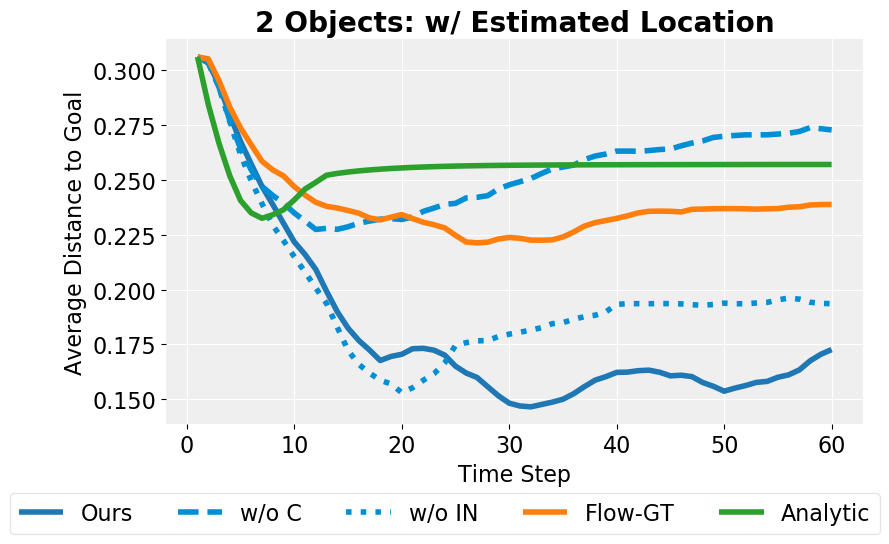} \hfill
    \includegraphics[height=1.35in]{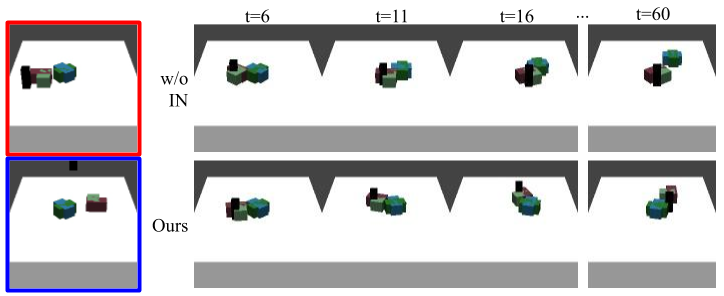}
    \caption{Left: Quantitative result in  test subset requiring objects to be pushed around the other. Right: Visualizing an executed action sequence in harder test subset in simulation. The blue block need to be pushed around the other object to achieve the goal. }
    \label{fig:exp:hard}
\end{figure}
\noindent \textbf{How important is the interaction?} We replace the interaction network with a simpler CNN that models independently the effects of action for each object i.e. no interaction. We create a harder dataset in simulation where one block is in the middle of the way for the other block reaching the goal. In this setting, it is more crucial to understand interaction/collision. Figure \ref{fig:exp:hard}  reports our model, the ablative model (w/o IN) in comparison of two strong baselines. Without IN, the performance is slightly better than our full model at the beginning but degrades more after $T=30$. This is probably because the model without interaction is more greedy i.e. makes progress initially, but fails to pass around objects.  
% Analytic baseline greedily select the object most further from goal location and push towards the goal. 
The analytic baseline performs much worse because the simple dynamic cannot estimate the location well since collisions will happen. Figure \ref{fig:exp:hard} shows an example of executed actions. Our model can push the object around the other  object because it learns a good model of interaction among objects.

\begin{figure}
    \centering
    \vspace{-1em}
    \includegraphics[width=0.37\linewidth]{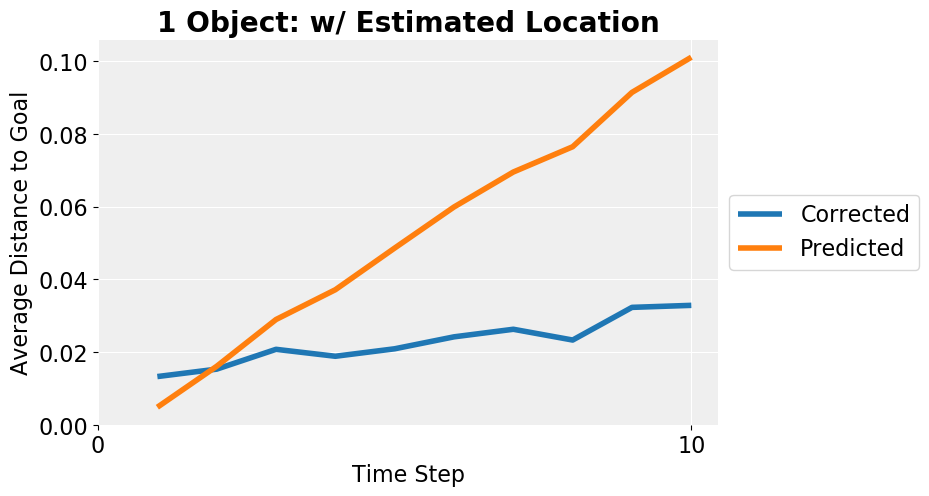} \hfill
    \includegraphics[width=0.6\linewidth]{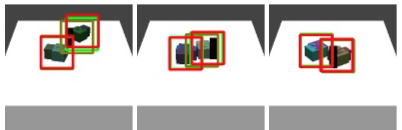}
    \caption{Left: Prediction error with respect to ground truth with or without correction. Right: Visualization of correction model. Ground truth is plotted green; the predicted location from forward model is plotted brown; the corrected one is plotted red.}
    % \vspace{-3mm}
    \label{fig:exp: correction}
\end{figure}

\noindent \textbf{Ablating Correction model.} We ablate the effect of the correction model using two metrics. First, in Figure \ref{fig:exp:hard}, we evaluate our model in MPC setting. `w/o C’ estimates location with predicted output without correction and it performs poorly without correction model to close the loop. Secondly, we evaluate it in terms of reducing the prediction error. In Figure \ref{fig:exp: correction} (Left), we measure the error between the predicted location and true location, when a 10-step prediction is unrolled with and without the correction module (when using correction module, we use intermediate observation to refine predictions). We see that the prediction error accumulates without any correction. 

Lastly, Figure \ref{fig:exp: correction} (Right) visualizes some qualitative results. A box around the ground-truth location is plotted as green; the predicted location output by the forward model is plotted as brown; the corrected location is plotted in red. Our model learns to correct the location when the prediction is inaccurate, and retain the predictions when accurate.

\begin{figure}
    \centering
    \includegraphics[width=.95\linewidth]{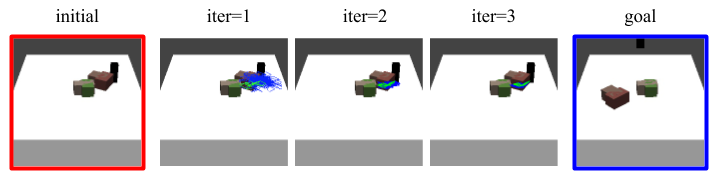}
    \caption{Visualization of sampled trajectories at different iteration of CEM. Trajectories are plotted in blue; elite samples are plotted in green.}
    % \vspace{-2mm}
    \label{fig:exp:cem}
\end{figure}
\noindent \textbf{Visualizing Planned Action Sequence.} We visualize in Figure \ref{fig:exp:cem} the $S=200$ action sequences sampled from the evolving Gaussian distribution across different iterations of the cross-entropy method (CEM) and highlight the $k=10$ best samples.
%At every iteration, based on these top samples, a new Gaussian distribution is estimated, using which $S$ new samples are again generated.
In the example depicted, we see that after several iterations the model converges to a non-greedy trajectory with the awareness of other objects.

\noindent \textbf{Visualizing Predictions.}
\begin{figure}[t!]
    \centering
    \includegraphics[width=\linewidth]{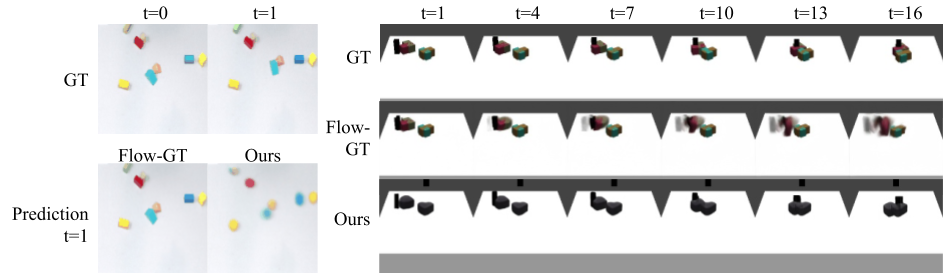}
    \caption{Visualizing prediction of forward model which unrolls the current observation for $T$ steps.}
    % \vspace{-1.5em}
    \label{fig:exp:pred}
\end{figure}
Figure \ref{fig:exp:pred} visualizes the prediction of forward model given the initial configuration $x^0$, and sequence of one or more actions $a^{1:T}$. In the synthetic data where only small motion happens, both our method and the baseline generate reasonable predictions. However, in the real dataset, the flow baseline cannot learn to predict the flow because the motion is relatively large. In contrast, in the predicted result of our model, when the blue one in the middle is pushed right, the orange one next to it also moves right due to interaction among them. 

% \subsection{Failure Cases}
% There are two typical failure modes. Firstly, when the 

\section{Conclusion}
We presented an object-centric forward modeling approach for model predictive control. By leveraging the fact that a scene is comprised of a collection of distinct objects, where each object can be described via its location and visual descriptor,  we designed a corresponding forward model that learns to predict in this structured space. We showed that this explicit structured representation better captures the interaction among objects and the robot, and thereby allows better planning in conjunction with an additional correction module. While we could successfully apply our system in both synthetic and real-world settings, we relied on explicit supervision on the object locations during training. It will be an interesting direction to further relax this assumption and let the objects emerge from unsupervised videos. Lastly, while we only modeled the effects of a single class of actions \ie pushing, it would be useful to generalize such prediction to work across diverse actions.

% The maximum paper length is 8 pages excluding references and acknowledgements, and 10 pages including references and acknowledgements

% The acknowledgments are automatically included only in the final version of the paper.
\vspace{2mm}
\textbf{Acknowledgements.} We thank the reviewers for constructive comments. Yufei would like to thank Tao Chen and Lerrel Pinto for fruitful discussion. This work was partially supported by MURI N000141612007 awarded to CMU and a Young Investigator award to AG.

%===============================================================================

\bibliography{ref}  % .bib

\newpage
\begin{appendices}
\section{Real Robot pushing}
\begin{figure}
    \centering
    \includegraphics[width=\linewidth]{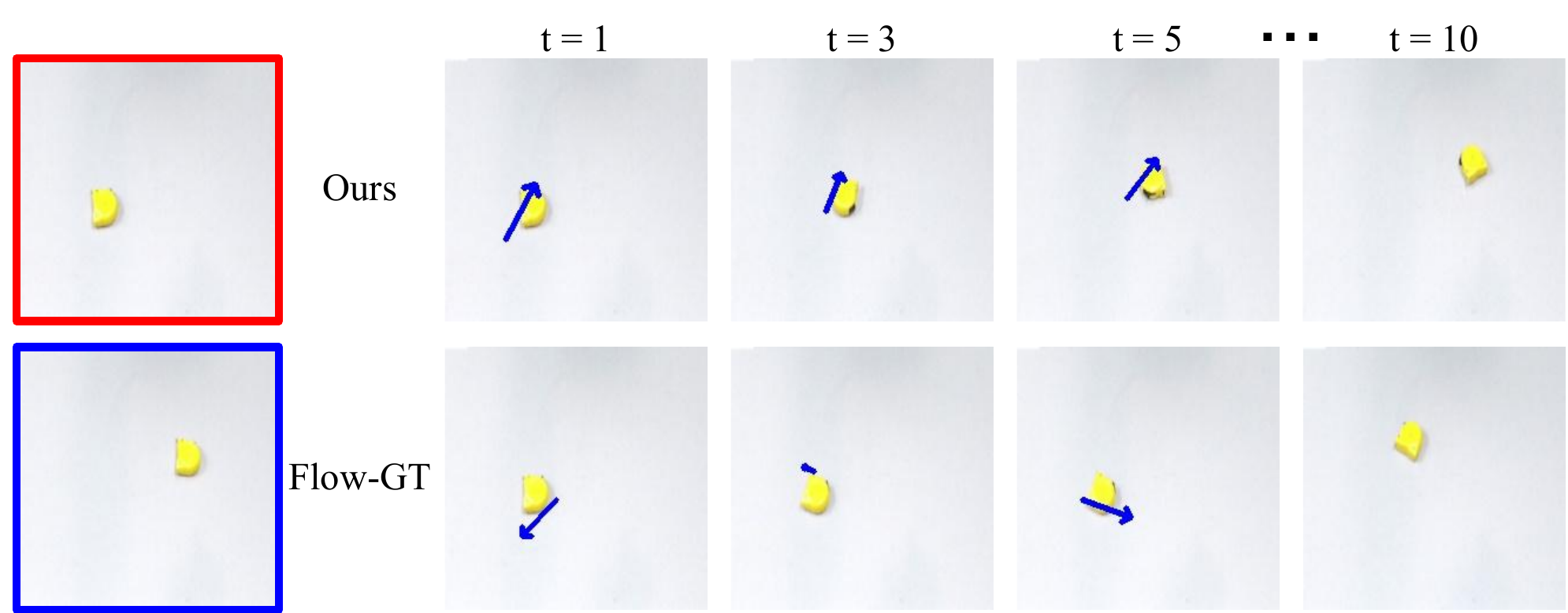}
    \caption{Blue arrow line shows the sequence of action taken by the robot to move objects from start configuration, shown in the red box, to a goal configuration, shown in the blue box.}
    \label{fig:exp:robot_novel_obj}
\end{figure}

\begin{figure}
    \centering
    \includegraphics[width=\linewidth]{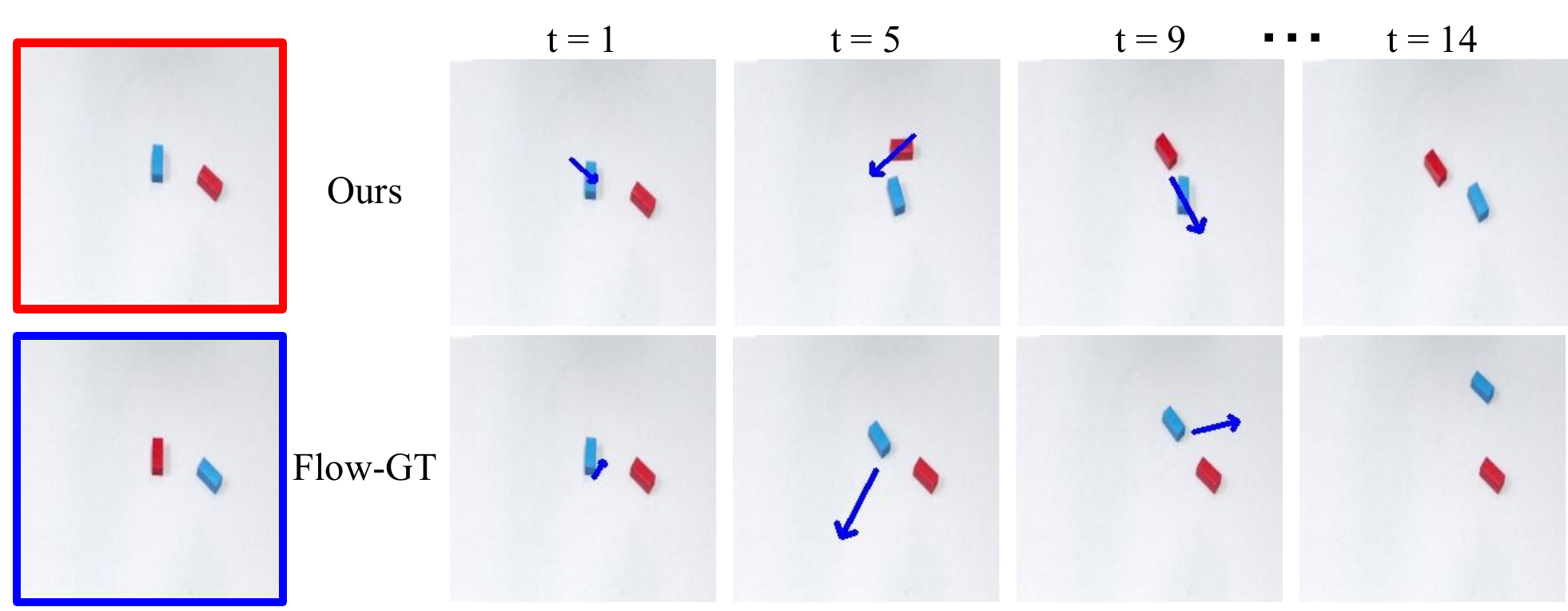}
    \caption{Blue arrow line shows the sequence of action taken by the robot to move objects from start configuration, shown in the red box, to a goal configuration, shown in the blue box.}
    \label{fig:exp:robot_multi_obejct_test_2}
\end{figure}

\begin{itemize}
    \item Robot Setup:To collect real world data we use Sawyer robot. We place a table in front of it where the objects are placed in order for robot to push it. Kinect V2 camera is rigidly attached overlooking the table for RGB-D perception data. The  camera is  localized  with  respect  to  the  robot  base  via calibration  procedure.
    \item Data Collection Procedure: Given the image $I_{s}$ of table  with object on it, we first perform the background subtraction to get the binary mask corresponding to objects . Using this binary mask, we sample a pixel $P_{m}^{C}$ which lies on the object. We treat $P_{m}^{C}$ as the mid-point of push. For push start pixel $P_{s}^{C}$, we sample pixel around $P_{m}^{C}$ in square such a way that it does not lie on top of the object. The end point of the push $P_{e}^{C}$ is calculated based on $P_{s}^{C}$ and $P_{e}^{C}$. These pixel location $P_{s}^{C}, P_{e}^{C}$ in image space are converted to corresponding 3D points $P_{s}^{R}, P_{e}^{R}$ in robot space using the depth image and camera matrix. Then we use off-the-shelf-planner to move robot gripper finger from $P_{s}^{R} \rightarrow P_{e}^{R}$. The image $I_{e}$ is recorded after the arm retracts back. For every push we record the tuple of $(I_{s}, I_{e}, P_{s}^{R}, P_{e}^{R})$. Figure \ref{fig:eral_data} shows some of the pushing data point collected on real robot. In all we have collected 5K pushing data-points on 8 objects.
    \item Push novel object: To see how well our method generalizes to novel object, we tested it out for pushing measuring tape. In figure \ref{fig:exp:robot_novel_obj} blue arrow shows the push predicted by our method to move it to desired location. Even though our method hasn't seen this object during training of forward model, it is able to push it very close to goal location.
    \item Flip the object location: To test the effectiveness of our method, we tested it on a bit more challenging scenario. In this case, we have 2 objects on the table. The goal configuration is generated by interchanging the position of objects in start configuration. Figure \ref{fig:exp:robot_multi_obejct_test_2} shows the sequence of action taken by our method to carry out this task.  
\end{itemize}

\section{Baseline Model Details.}
\begin{itemize}
    \item Implicit forward model (\textbf{Imp-Inv})  \cite{agrawal2016learning}: the model predicts in a implicit feature space where the entire frame is encoded as one implicit feature without further factoring to objects.  An inverse model is trained to take in current and goal feature and outputs the action. In testing, the inverse model are applied iteratively to greedily generate action sequence. The inverse model also regularize the forward model to prevent trivial solution.
    \item Implicit forward model with pixel reconstruction (\textbf{Imp-Plan}): The baseline is a variant of Imp-Inv. The action sequences are generated by a planner in the learned feature space. To further regularize the forward model such that it learns a more informative feature space, we train an decoder to reconstruct the frame in pixels. The learned representation of the frame is used by the planner.
    \item Flow-based prediction model SNA \cite{ebert2017self} (\textbf{Flow}): the model learns to predict transformation kernel to reconstruct future frame.  In planning, the predicted transformations are applied to designated pixels (location) to estimate  their motion.
    \item Flow baseline with supervision (\textbf{Flow-GT}): The original flow baseline only trains with videos  in the unsupervised manner. To leverage the additional information, we provide its variant -- besides transforming the pixels, the model also transforms the ground truth location $\{\hat b_n^t\}$ to $\{b_n^{t+1}\}$ and  minimizes the expected distance of transformed location to the ground truth $\{\hat b_n^{t+1}\}$. 
    \item \textbf{Analytic} baseline. To leverage the location information, a simple analytic solution is to greedily push in the direction of current goal position to desired position. It assume a simple dynamic -- the predicted location at the next step is calculated as the delta position of the gripper. 

\end{itemize}

\section{Plan with Oracle Location}
\begin{figure}
    \centering
    \includegraphics[height=1.5in]{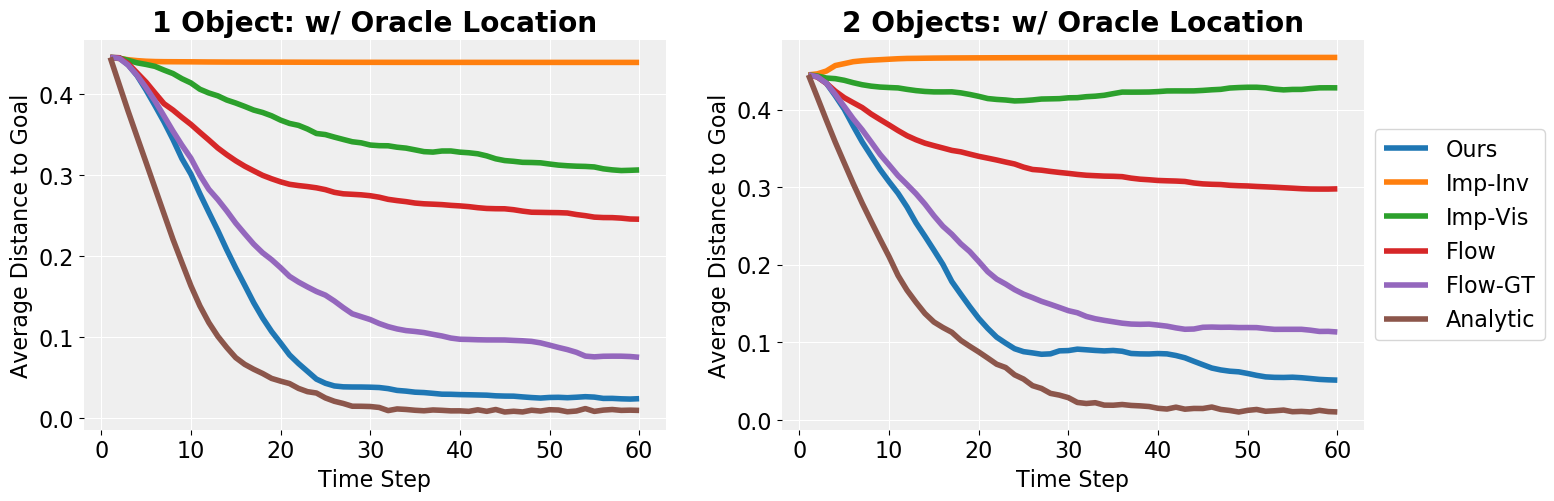}
    \caption{Distance to goal with access to ground truth location at every time step.}
    \label{fig:exp:ora}
\end{figure}
In this part we compare models when we have access to the ground truth location for each new observation at every time step. After every push, the distance between the current location and the goal in world coordinate is plotted in Figure \ref{fig:exp:ora}. The analytic baseline should converge to zero because the exact center of mass is given by oracle at every time step, hence serves as ceiling performance. The Imp-Inv barely generalizes to scenarios when the distance of goal and current observation is much ($\times 15$) farther than that in training set. Imp-Plan  degenerates in 2 blocks settings, suggesting one feature for the whole frame cannot encode complicated scenes very well. 
Flow works better than Imp-Plan because the motion space is more tractable. Its performance improves in Flow-GT to leverage location information. Our model outperforms other learning-based methods and performs comparably to the ceiling performance (Analytic) without manually specifying pushing toward goal through mass center.

\section{Qualitative Results of All Baselines.}
In this part we show qualitative results in comparison of all baselines. This supplements Figure \ref{fig:exp:est_vis}, which only showcases strong but partial baselines. For more results, please refer to project page.
\begin{figure}
    \centering
    \includegraphics[width=\linewidth]{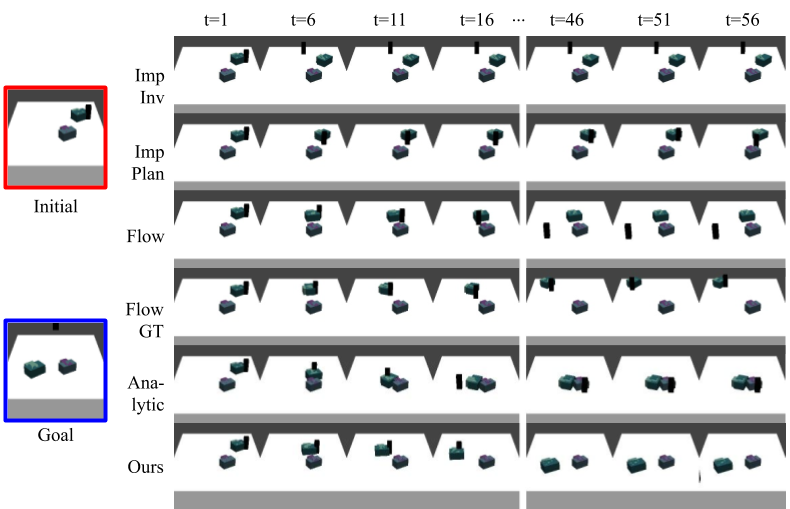}
    \caption{Visualizing an executed action sequence in simulation: Given the initial configuration (in red box) and the goal configuration (in blue box), figure shows the effect of the action predicted by various methods at different time steps. Please refer to the appendix to view all baselines.}
    \vspace{-1.5em}
    \label{fig:exp:est_full}
\end{figure}
\end{appendices}
\end{document}